\documentclass[conference]{IEEEtran}
\IEEEoverridecommandlockouts

\usepackage{cite}
\usepackage{amsmath,amssymb,amsfonts}
\usepackage{algorithmic}
\usepackage{graphicx}
\usepackage{textcomp}
\usepackage{xcolor}
\usepackage{tikz, pgfplots}
\usepackage{colortbl}
\usepackage{adjustbox}
\usepackage{balance}
\usepackage{siunitx}
\usepackage{enumitem}
\usepackage{afterpage}
\usepackage{caption}       
\usepackage{subcaption}    
\usepackage{placeins}
\usepackage{float}

\usetikzlibrary{shapes.arrows}
\def\BibTeX{{\rm B\kern-.05em{\sc i\kern-.025em b}\kern-.08em
    T\kern-.1667em\lower.7ex\hbox{E}\kern-.125emX}}

\makeatletter
\newcommand{\linebreakand}{%
  \end{@IEEEauthorhalign}
  \hfill\mbox{}\par
  \mbox{}\hfill\begin{@IEEEauthorhalign}
}

\usepackage{diagbox}
\usepackage{booktabs}
    
\begin{document}

\title{Effects of Robot Competency and Motion Legibility on Human Correction Feedback}


\author{\IEEEauthorblockN{Shuangge Wang}
\IEEEauthorblockA{\textit{Yale University}\\
New Haven, CT, USA \\
shuangge.wang@yale.edu}
\and
\IEEEauthorblockN{Anjiabei Wang}
\IEEEauthorblockA{\textit{Yale University}\\
New Haven, CT, USA \\
anjiabei.wang@yale.edu}
\and
\IEEEauthorblockN{Sofiya Goncharova}
\IEEEauthorblockA{\textit{Phillips Exeter Academy}\\
Exeter, NH, USA \\
sgoncharova@exeter.edu}
\linebreakand
\IEEEauthorblockN{Brian Scassellati}
\IEEEauthorblockA{\textit{Yale University}\\
New Haven, CT, USA \\
brian.scassellati@yale.edu}
\and
\IEEEauthorblockN{Tesca Fitzgerald}
\IEEEauthorblockA{\textit{Yale University}\\
New Haven, CT, USA \\
tesca.fitzgerald@yale.edu}}


\maketitle

\begin{abstract}

As robot deployments become more commonplace, people are likely to take on the role of supervising robots (i.e., correcting their mistakes) rather than directly teaching them. Prior works on Learning from Corrections (LfC) have relied on three key assumptions to interpret human feedback: (1) people correct the robot only when there is significant task objective divergence; (2) people can accurately predict if a correction is necessary; and (3) people trade off precision and physical effort when giving corrections. In this work, we study how two key factors (robot competency and motion legibility) affect how people provide correction feedback and their implications on these existing assumptions. We conduct a user study ($N=60$) under an LfC setting where participants supervise and correct a robot performing pick-and-place tasks. We find that people are more sensitive to suboptimal behavior by a highly competent robot compared to an incompetent robot when the motions are legible ($p=0.0015$) and predictable ($p=0.0055$). In addition, people also tend to withhold necessary corrections ($p < 0.0001$) when supervising an incompetent robot and are more prone to offering unnecessary ones ($p = 0.0171$) when supervising a highly competent robot. We also find that physical effort positively correlates with correction precision, providing empirical evidence to support this common assumption. We also find that this correlation is significantly weaker for an incompetent robot with legible motions than an incompetent robot with predictable motions ($p = 0.0075$). Our findings offer insights for accounting for competency and legibility when designing robot interaction behaviors and learning task objectives from corrections.
\end{abstract}




\maketitle

\section{Introduction}
\label{section: introduction}


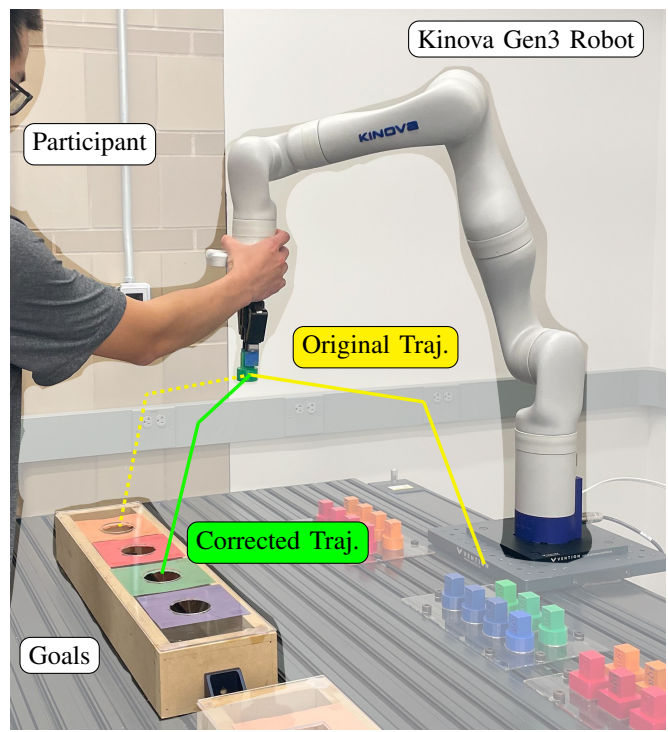
\begin{figure}
\centering
    \begin{tikzpicture}
        \node[inner sep=0pt] (img) {\includegraphics[width=0.48\textwidth, trim = 1400 1200 500 1300, clip]{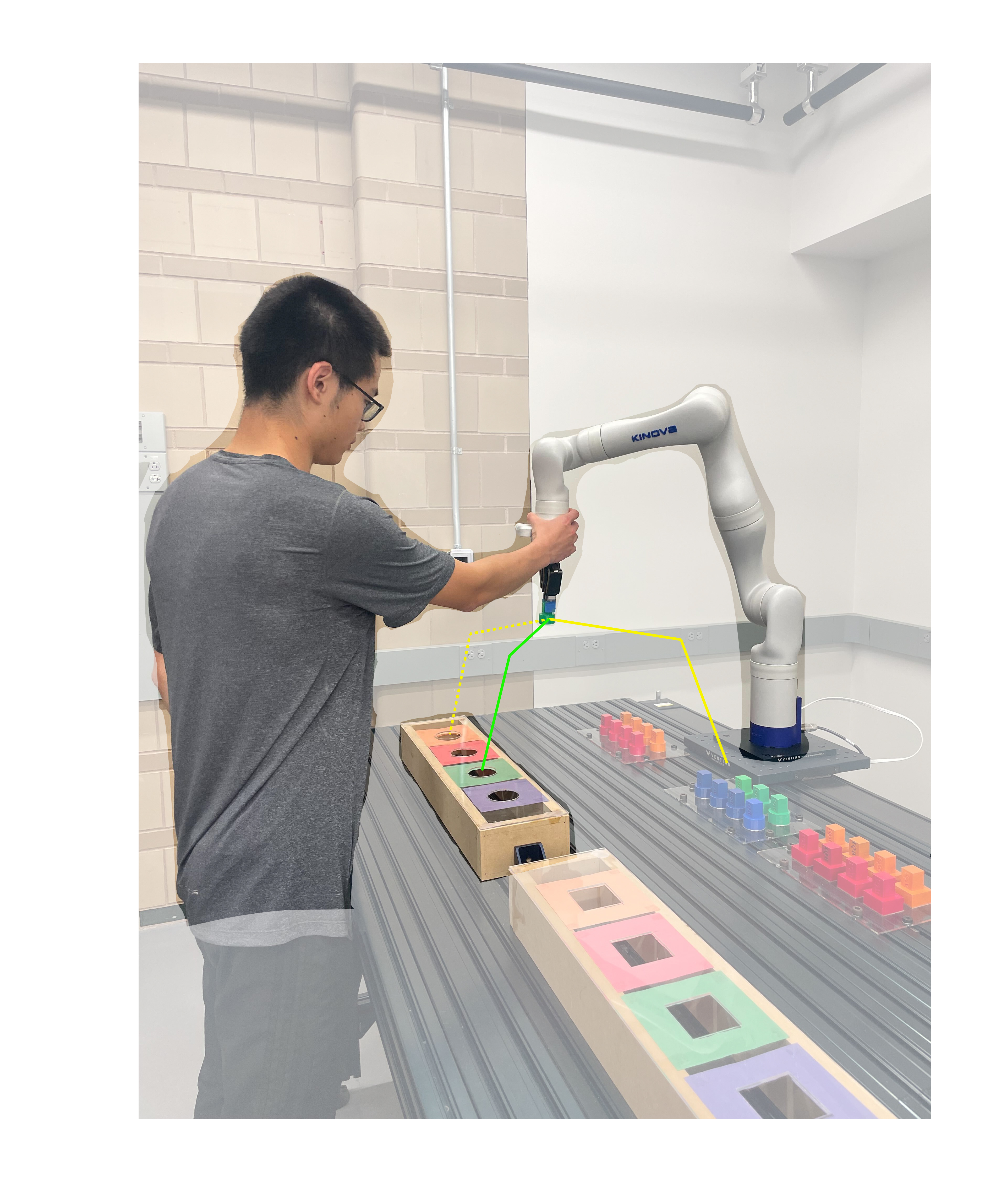}};
        \node[draw, rectangle, rounded corners, fill=white, align=center] at (2.5, 4.4) {Kinova Gen3 Robot};
        \node[draw, rectangle, rounded corners, fill=white, align=center] at (-3.3,3) {Participant};
        \node[draw, rectangle, rounded corners, fill=white, align=center] at (-3.7,-3.8) {Goals};
        \node[draw, rectangle, rounded corners, fill=yellow, align=center] at (0.5,0.3) {Original Traj.};
        \node[draw, rectangle, rounded corners, fill=green, align=center] at (-0.8,-2.3) {Corrected Traj.};
    \end{tikzpicture}
    \caption{The robot begins moving along the yellow trajectory to place a shape in one of the target holes. The participant intervenes, guiding the robot to a different target via the green trajectory, leaving the dashed portion of the robot's intended trajectory untraveled.}
    \vspace*{-5mm}
    \label{fig:correction}
\end{figure}


Imagine that you are developing robots that will be deployed in warehouses and tasked with packing merchandise for shipment. These robots are pre-trained with basic abilities such as recognizing boxes and manipulating a range of items. Despite this pre-deployment training, they will inevitably make mistakes with identifying and manipulating novel inventory that is specific to the warehouse where the robot is deployed. Furthermore, new inventory may be introduced over time. In order for robots to function effectively post-deployment, they should learn continuously from their human co-workers and collaborators as novel objects and task constraints arise.


Traditionally, robots learn from people during training sessions that are distinct from its deployment/test sessions. In these training sessions, a person directly teaches the robot by providing numerous demonstrations of the desired behavior~\cite{atkeson1997robot, ekvall2008robot, schaal1996learning, rozo2013robot, arduengo2023gaussian} or indicating preferences between numerous hypothetical robot behaviors~\cite{akrour2011preference, akrour2012april,munzer2017preference, wilde2020improving, wang2022skill, hejna2023few}.

There has been comparatively less work on how people might teach robots by supervising its behavior \emph{during} deployment~\cite{faulkner2018policy, mcpherson2018modeling, schreckenghost2008human, lang2003providing}.
In a supervisory paradigm, a person observes the robot's behavior as it attempts to complete a task, intervening only when necessary to avoid an impending mistake or to train the robot to avoid future mistakes~\cite{bajcsy2017learning, bajcsy2018learning, losey2018including, bobu2020less, bobu2020quantifying, losey2022physical, chisari2022correct, wang2024legibility, fitzgerald2021modeling, fitzgerald2019human, jevtic2018robot, spencer2020learning, spencer2022expert, liu2023interactive, wang2024toward}. Compared to traditional training sessions, this paradigm leverages the robot's autonomy whenever possible and avoids the need for continuous teaching effort by the human during long-horizon tasks~\cite{sheridan1986human}. However, supervisory feedback might not be as abundant~\cite{mohtasib2021study}; the human might not be as attentive~\cite{lemaignan2016real, chen2020robot}; and the quality of human feedback might not be as consistent~\cite{valletta2021imitation, qian2023robot}. 

Yet, current learning techniques do not account for these differences when learning from supervisory feedback, which may impair the robot's ability to learn accurate and precise task objectives.
%
%
%
Existing learning algorithms are designed under the assumption that (1) people intervene and correct the robot only when it is about to make a mistake; (2) people can accurately predict when they do or don't need to intervene; and (3) people trade off precision and physical effort when providing corrections. 
%
Prior works have yet to explore whether these assumptions are valid, or what other factors may influence how people correct a robot during long-duration tasks. From a learning perspective, it is important to identify and factor out these influences to facilitate effective learning for the underlying task information. Furthermore, to design robots and algorithms that people will adopt into their homes and workplaces, we need to understand better how a person reacts to these factors when supervising a robot.

Our work aims to study how people give feedback to a robot in a supervisory setting.
We study two factors that we expect are particularly important in influencing how people correct the robot: robot competency (i.e., the robot's prior task performance) and motion legibility (i.e., the interpretability of the robot's motion toward its goal).
We present an between-subject user study ($N=60$) where we examine how the robot's competency and legibility conditions influence how participants supervise it through a series of 64 pick-and-place actions (Fig.~\ref{fig:correction}). 
To the best of our knowledge, our work is the first to identify how these features of a robot's behavior affect people's correction feedback as they supervise a robot. 
We provide the following contributions:
\begin{itemize}
\item We design a learning from corrections study to measure how people supervise a robot's behavior differently depending on its competency and motion legibility.
\item We study how these factors influence the timing and perceived necessity of corrections.
\item We examine the trade-off between precision and physical effort as people provide corrections.
\item We propose recommendations for (1) designing robots to elicit better correction feedback post-deployment and (2) interpreting this feedback as training data.
\end{itemize}


\section{Related Work}
\label{section: related work}


\subsection{Learning from Corrections}
\label{subsection: learning from corrections}

Research in interactive robot learning has explored multiple modalities through which people can provide training data and feedback to a robot, such as rewards~\cite{knox2009interactively, daniel2014active, chernova2014robot}, demonstrations~\cite{schaal1996learning, atkeson1997robot, ekvall2008robot, rozo2013robot, arduengo2023gaussian}, preferences~\cite{akrour2011preference, akrour2012april,munzer2017preference, wilde2020improving, wang2022skill, hejna2023few}, physical corrections~\cite{bajcsy2017learning, bajcsy2018learning, losey2018including, bobu2020less, bobu2020quantifying, losey2022physical, chisari2022correct, wang2024legibility, fitzgerald2021modeling, fitzgerald2019human}, implicit feedback~\cite{cui2021empathic, loftin2014learning, loftin2016learning}, and natural language~\cite{lauria2001personal,matuszek2013learning, thomason2015learning, stepputtis2020language}. These modalities differ in the role that they prescribe to the human and the robot during the interaction~\cite{cui2021understanding}. Some modalities position humans in a more supervisory role than others, such as interventions~\cite{jevtic2018robot, spencer2020learning, spencer2022expert, liu2023interactive}, negative reinforcement~\cite{thomaz2006reinforcement, thomaz2008teachable, navarro2017improving}, implicit feedback~\cite{cui2021empathic, loftin2014learning, loftin2016learning}, and physical corrections~\cite{bajcsy2017learning, bajcsy2018learning, losey2018including, bobu2020less, bobu2020quantifying, losey2022physical, wang2024legibility,fitzgerald2019human,fitzgerald2021modeling}.

Correction feedback, in particular, involves a robot attempting to complete a task while supervised by a human teacher. The teacher can intervene and modify the robot's motion kinesthetically, producing a \emph{corrected} trajectory that is assumed to be more optimal with respect to the hidden task objectives~\cite{bajcsy2017learning, bajcsy2018learning, losey2018including, bobu2020less, bobu2020quantifying, losey2022physical, wang2024legibility,fitzgerald2019human,fitzgerald2021modeling}. While prior works have proposed methods for Learning from Corrections (LfC), they rely on three important assumptions.

\textbf{Assumption 1: People Correct Only When There Is Significant Task Objective Divergence}

Prior works in learning from interventions~\cite{liu2023interactive, spencer2020learning, spencer2022expert} and corrections~\cite{chisari2022correct, wang2024legibility} assume that people decide to intervene and correct the robot only when the robot is about to make a mistake. Quantitatively, we can represent this decision as a threshold for allowable divergence between the robot's behavior and the ``correct" goal for its task. As an example, a robot that moves away from an object it should be picking up and toward objects that it should be avoiding is \emph{increasing} this divergence. Prior work assumes that this threshold is either consistent across all users~\cite{fitzgerald2022inquire}, is user-specific~\cite{ramachandran2007bayesian, carreno2022joint}, or task-specific~\cite{ziebart2008maximum}, but does not consider how the robot's own behavior might influence this threshold.

\textbf{Assumption 2: People Can Accurately Predict If a Correction is Necessary}

A person's decision to correct a robot's motion can be viewed as a label for incorrect behavior; conversely, the lack of an intervention can be considered to be an endorsement~\cite{thomaz2006reinforcement, thomaz2008teachable}. Current methods for learning from human-provided labels require that the human achieves optimal or near-optimal labelling accuracy~\cite{nourani2019effects, tecimer2021detection}. While previous works have attempted to design algorithms that learn from noisy human-provided labels~\cite{gupta2018robot, wei2021learning}, the robot's learning performance will still be affected by inaccurate labels.

\textbf{Assumption 3: People Trade Off Precision and Effort}

Prior works in interactive robot learning assume that the teacher is incentive-driven~\cite{ziebart2008maximum, ramachandran2007bayesian, finn2016guided, bloem2014infinite, dragan2013legibility}, providing feedback to maximize perceived reward and minimize perceived cost. Existing LfC models, in particular, assume that people aim to provide corrections that optimize task performance, while being biased toward corrections that require less physical effort to provide~\cite{bajcsy2017learning, bajcsy2018learning, losey2018including, bobu2020less, bobu2020quantifying, losey2022physical, wang2024legibility}.
Yet, no prior work has empirically validated this trade-off. 

\subsection{Robot Competency}
\label{subsec: influences}

In response to these assumptions, we now consider additional factors that may influence how people correct a robot's behavior. Deployed robots will naturally exhibit different levels of competency across various tasks based on how well their training data aligns with their deployment environment~\cite{zhao2020sim} or constraints of its sensor, actuator, and computational capabilities~\cite{afzal2020study,neuman2022tiny}. 
Prior works have shown that the robot's competency is one of the most prominent factors in shaping people's trust~\cite{hancock2011meta}, expectation~\cite{paepcke2010judging}, and preferences~\cite{scheunemann2020warmth} over the robot's behaviors. DelPreto et al.~\cite{delpreto2020helping} found that when a robot performed a task with lower accuracy, participants’ trust in the robot decreased, their perception of the robot's intelligence decreased, and their
workload increased. Hedlund et al.~\cite{hedlund2021effects} showed that an incompetent robot led people to lower their trust in both the robot and in their own teaching ability. Paepcke et al.~\cite{paepcke2010judging} found that setting low expectations for the robot's ability at a task resulted in less disappointment and more positive appraisals from participants.

Overall, these prior works study how the robot's competency affects a person's subjective perception of it. Yet, they do not evaluate how this subjective perception influences how the person provides feedback to the robot. 
From a robot learning perspective, other works have focused on the effect of robot competency in a setting where people serve as direct teachers (i.e., in learning from demonstration)~\cite{delpreto2020helping, hedlund2021effects}. Yet, this remains unexplored in \emph{supervisory} paradigms such as LfC.


\begin{figure}
    \centering
    \includegraphics[width=0.8\linewidth]{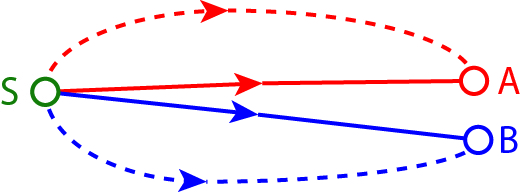}
    \caption{A comparison between predictable (solid lines) and legible motions (dashed lines) starting from S to two potential goals, A and B.}
    \vspace*{-5mm}
\label{fig:legibility}
\end{figure}

\subsection{Motion Legibility}
\label{subsec: legibility}

While competency describes the robot's \emph{past} behavior, we now consider how the robot's \emph{current} behavior (i.e., its motion trajectory) influences a person's perception of it while they supervise the robot. Typical motion planning algorithms optimize for short and efficient trajectories between the robot's starting position and its goal position. This produces a trajectory that is \emph{predictable}; i.e., the motion that a person would expect to see if they \emph{already know} what goal position the robot is trying to reach~\cite{dragan2013legibility}. To a person who does \emph{not} know the robot's goal, however, it may be difficult for them predict the robot's goal in real-time as the robot moves. 
The aim of \emph{legible} motion is to enable an observer to quickly predict the robot's goal from its early motion~\cite{dragan2013generating,dragan2013legibility}. This typically results in the robot exaggerating its motion toward its goal, as shown in Fig.~\ref{fig:legibility}.

Prior works have verified that legible motion is more effective at conveying the robot's intent to a human observer~\cite{dragan2013legibility, dragan2013generating, stulp2015facilitating} and can elicit more informative feedback~\cite{sadigh2017active, hu2022active, wang2023active}. Furthermore, legible motion can improve a person's perceived safety~\cite{lichtenthaler2012influence}, comfort~\cite{hetherington2021hey}, trust~\cite{dragan2015effects}, and positive affect of robots~\cite{kim2017review}. 
While prior works have focused on generating legible motions~\cite{dragan2013generating, dragan2013legibility} and incorporating legibility while modeling human feedback in task learning~\cite{wang2024legibility, busch2017learning}, its nuanced effects on the nature and quality of human feedback remain largely unexplored.

\section{Research Questions} 
\label{subsubsection:research questions}

Our work investigates the influence of the robot's competency and motion legibility on how people correct its behavior. Our aim is to inform how we design (1) interactions for robots to elicit better correction feedback and (2) learning algorithms to interpret corrections for more accurate task models.
We focus on the following three research questions (RQs) investigating how competency and legibility affect the validity of the assumptions described in Section~\ref{subsection: learning from corrections}. 

\subsection*{RQ1: How do competency and legibility affect when people correct the robot?}

%
In response to \textbf{Assumption 1} (Section~\ref{subsection: learning from corrections}), we consider how competency and legibility may aggravate or attenuate a person's threshold for divergence (which informs their decision of when to correct the robot).
%
We expect that a robot exhibiting legible motion will enable people to infer the goal of its motion earlier in its trajectory execution, allowing them to more quickly assess whether it will be aligned with the correct goal (and thus, whether a correction is necessary). We also expect that people will trust a highly competent robot more, thus raising their threshold for divergence and resulting in later corrections. We establish the following hypotheses:

\noindent
\textbf{H1A}: When supervising robots with legible motion, people will correct the robot earlier in the trajectory (i.e., when there is a smaller task objective divergence).

\noindent
\textbf{H1B}: When supervising a highly competent robot, people will correct the robot later in the trajectory (i.e., will have a higher tolerance for task objective divergence).

\subsection*{RQ2: How do competency and legibility affect people's accuracy in predicting robot success/failure?}

In response to \textbf{Assumption 2}, we expect that people will distrust an incompetent robot, and thus are more likely to predict that it will fail (even if it would have succeeded using its intended trajectory). Similarly, we expect that people will trust a highly competent robot, and thus are more likely to predict that a failing robot will eventually succeed (until it is too late for them to provide a correction). Since legible motion enables a person to predict the robot's goal sooner, we also expect legible motions to improve predication accuracy. We establish the following hypotheses:

\noindent
\textbf{H2A}: People are more likely to miss necessary corrections in high-competency conditions.

\noindent
\textbf{H2B}: People are more likely to provide unnecessary corrections in low-competency conditions.

\noindent
\textbf{H2C}: Legible motions will increase prediction accuracy, regardless of competency condition.

\subsection*{RQ3: How do competency and legibility affect how people balance precision and effort in their corrections?}

In response to \textbf{Assumption 3}, we aim to confirm whether there is a consistent trade-off between the precision and physical effort that people expend as they correct the robot's motion.
\textbf{H3}: People's corrections will exhibit a trade-off between precision and physical effort.

\section{Methodology}
\label{sec:methodology}

We conducted a user study with 60 participants recruited from our university community. Participants reported their age ($M = 27.53$ years, $SD = 9.49$ years) and gender (32 male, 27 female, 1 non-binary). Each participant supervised a robot arm performing a series of pick-and-place tasks. The task goal was for the robot to place each shape into the target hole with the corresponding color. 
Participants were instructed to interrupt the robot's motion and provide a correction whenever and however they saw fit to guide the robot toward successfully completing the task (Fig.~\ref{fig:correction}).

To incentivize high-quality data (as would be expected of a person earnestly trying to train a robot collaborator), participants were told that the robot was learning from their feedback in real-time, and that they would receive additional compensation based on the number of successful robot trials. In reality, the robot followed pre-determined waypoints based on the participant's study condition (rather than learning in real-time), and participants received the maximum compensation (as if the robot had succeeded at every trial) to ensure that they were fairly compensated regardless of their study condition. We obtained approval for this study through our Institutional Review Board (IRB) and followed ethics protocol for debriefing participants on these hidden elements.


\subsection{Experimental Design}
\label{fig:experimental design}

Our experiment involved two independent variables: competency (consisting of two levels) and legibility (consisting of three levels). Our experiment was thus a between-subject 2$\times$3 user study, with 10 participants assigned to each condition. In each condition, participants supervised 64 task trials, divided into 4 sub-tasks (i.e., 4 different shapes) with 16 trials each.

Throughout this section, \emph{actual} success/failure refers to whether the robot places a shape in the correct target. \emph{Intended} success/failure refers to the result of the robot's planned trajectory without correction.


\subsubsection{Independent Variables}

\begin{figure}
    \centering
    \includegraphics[width=0.48\textwidth]{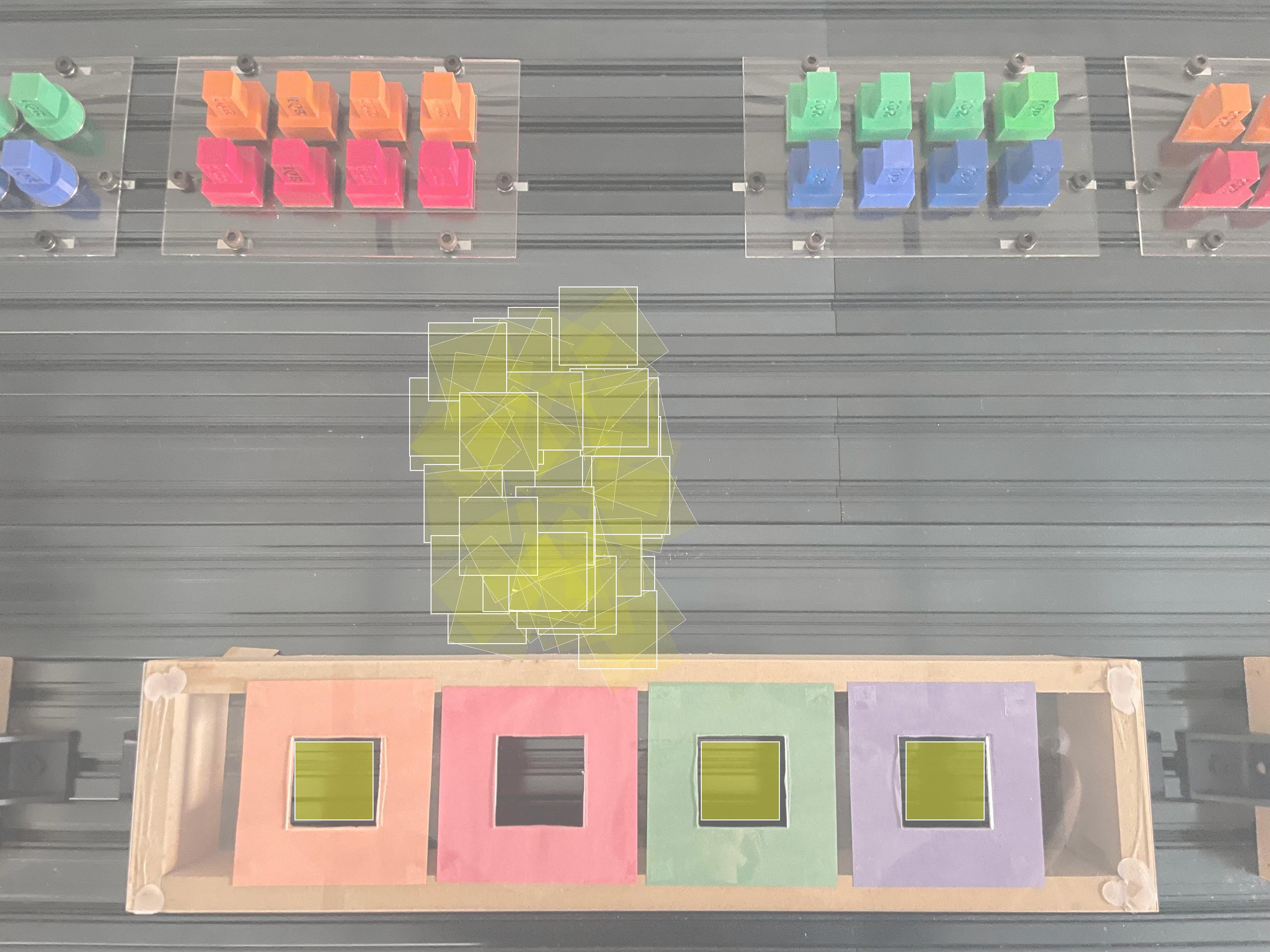}
    \caption{In an intended-success trial, the robot will attempt to place a red square shape into the red target. In an intended-failure trial, the robot will attempt to place the red shape into the wrong-colored target or onto the workbench surface. The yellow-shaded squares represent the distribution of possible locations where the robot may attempt to place the shape.}
    \label{fig:error_distribution}
    \vspace*{-5mm}
\end{figure}

In \textbf{low competency} conditions, the robot intended to succeed in only 25\% of trials. Among the intended failures, 50\% involved the robot placing the shape in a hole of the wrong color. The other 50\% involved the robot missing the target holes entirely, with intended failure poses uniformly sampled from a neighborhood near the correct target. Fig.~\ref{fig:error_distribution} illustrates the combined probability distribution over potential goals for an intended failure.
%
In \textbf{high competency} conditions, the robot intended to succeed in 75\% of trials, with intended failures also following the distribution shown in Fig.~\ref{fig:error_distribution}. 
The intended failures were distributed across colors and trials to minimize bias. 

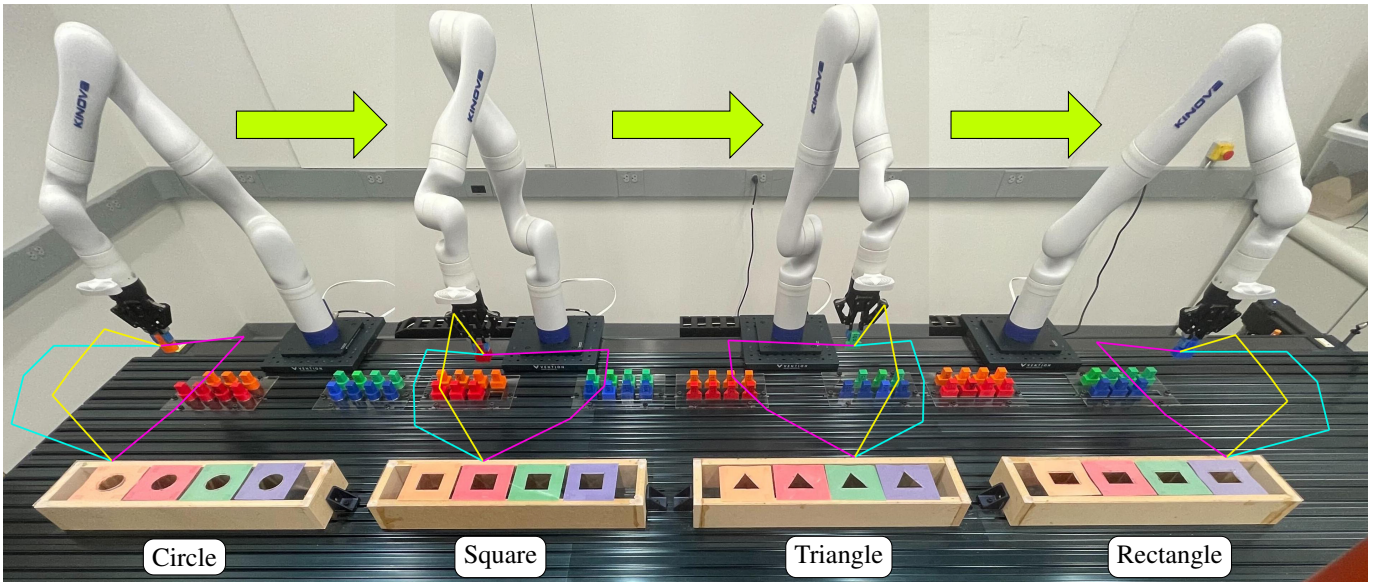
\begin{figure*}
    \includegraphics[width=\textwidth, trim = 00 150 00 00, clip]{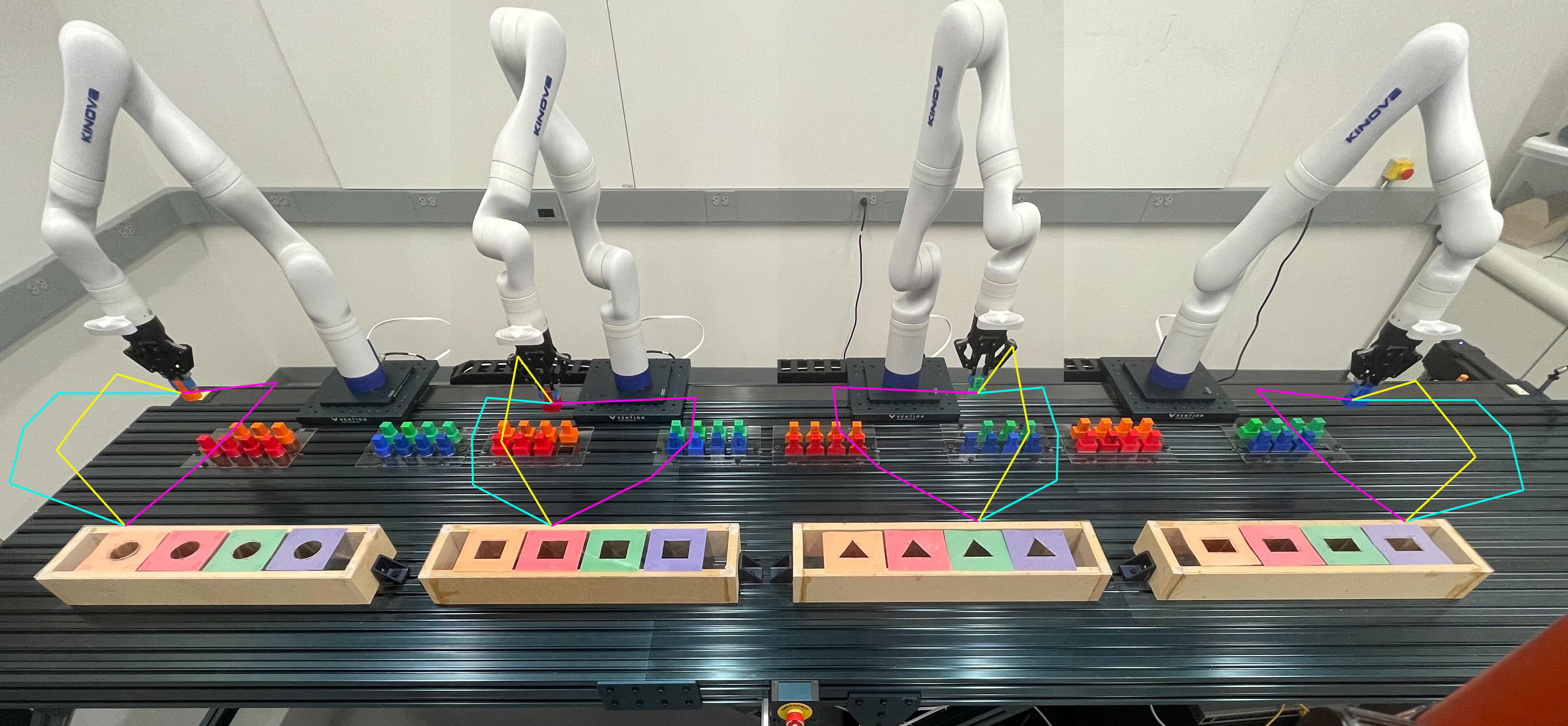}

    \begin{tikzpicture}[overlay, remember picture]
        \node[single arrow, draw=black, fill=lime, minimum width=25pt, 
              single arrow head extend=5pt, minimum height=20mm] 
              at (4.0, 6.5) {};  
    
        \node[single arrow, draw=black, fill=lime, minimum width=25pt, 
              single arrow head extend=5pt, minimum height=20mm] 
              at (9.0, 6.5) {};   
    
        \node[single arrow, draw=black, fill=lime, minimum width=25pt, 
              single arrow head extend=5pt, minimum height=20mm] 
              at (13.5, 6.5) {};   

        \node[draw, rectangle, rounded corners, fill=white, align=center] at (2.4, 0.75) {Circle};
        \node[draw, rectangle, rounded corners, fill=white, align=center] at (6.6, 0.75) {Square};
        \node[draw, rectangle, rounded corners, fill=white, align=center] at (11.1, 0.75) {Triangle};
        \node[draw, rectangle, rounded corners, fill=white, align=center] at (15.5, 0.75) {Rectangle};
    \end{tikzpicture}
    
    \caption{This figure depicts the series of pick-and-place tasks performed by the robot. The tasks involved manipulating various shapes in sequence: circle, square, triangle, and rectangle. Each shape had 4 colors: orange, red, green, and blue. The colored lines illustrate the robot's trajectories, categorized by their legibility: cyan represents legible paths, yellow represents predictable paths, and purple represents illegible paths.}
    \label{fig:setup}
    \vspace*{-5mm}
\end{figure*}

Motion legibility~\cite{dragan2013legibility} consisted of 3 levels, shown in Fig.~\ref{fig:setup}:
\begin{itemize}
\item \textbf{Predictable} motion was short and efficient; i.e., the default output of an RRT* motion planner.
\item \textbf{Legible} motion enabled the observer to quickly infer the robot's end-goal.
\item \textbf{Illegible} motion obscured the robot's end goal by initially moving toward goals that the robot did not plan to visit. 
\end{itemize}


\subsection{Robot Control \& Motion Planning}
\label{subsubsection:setup}

We performed the study using a Kinova Gen3 7-DoF robot arm equipped with a Robotiq 2F-85 gripper as its end effector (EEF). The arm was mounted on a horizontal linear actuator, allowing it to slide along the workbench to access each of the four sub-tasks. We computed Cartesian waypoints for every trajectory for each legibility level by optimizing their legibility score until convergence~\cite{dragan2013legibility}. We then used RRT* to plan the Cartesian waypoints into joint space trajectory~\cite{sucan2012the-open-motion-planning-library}. We down-sampled and interpolated the planned trajectory for improved efficiency and smoothness. The trajectory was then executed via a velocity-based PID controller running at sub-1000 Hertz. To avoid bias toward certain colors, we uniformly randomized the order in which the robot picked up shapes in each sub-task.

We implemented admittance (inverse impedance) control~\cite{kim2019model, jenamani2024feel} to enable the robot to quickly follow any physical force exerted by the participant. We used the recursive Newton-Euler algorithm to compute the inverse dynamics and gravity compensation~\cite{carpentier-rss18}. We collected joint encoder and torque sensor readings and control input at 10Hz and applied low pass filters to denoise the data. After the participant stopped applying force to the robot arm (i.e., after finishing the correction), the robot replanned and executed a trajectory from its new state to the nearest target goal, maintaining the EEF rotation as it was at the end of the correction. The participant may provide additional corrections during the robot's replanned motion; regardless, we focus our analyses on the \emph{first} correction within each trial. 

\subsection{Procedures}
\label{subsection: procedures}


\subsubsection{Pre-Interaction Phase}
\label{subsubsection: pre-interaction phase}

Participants were pre-screened to meet the criteria of being at least 18 years old, fluent in English, able to stand for up to an hour, and not color blind. After the screening, the participant consented to the experiment and video recording. The experimenter followed a script to familiarize the participant with the robot, the pick-and-place task, and their role in correcting the robot.

\subsubsection{Tutorial Phase}
\label{subsubsection: admittance control phase}

The researcher activated the robot's admittance mode to enable the participant to practice moving the robot around.
%
Once the participant was ready to proceed, the experimenter played two example behaviors on the robot: one successful placing trial (requiring no correction) and one unsuccessful trial (where the experimenter demonstrated how to intervene and correct the robot). The participant then administered two similar trials. 

\subsubsection{Main Experiment Phase}
After completing the tutorial, the experimenter left the participant alone to supervise and correct the robot for all 64 pick-and-place trials. The experimenter remained on the other side of a dividing curtain, monitoring the experiment for safety via a webcam and staying within reach of an emergency-stop button.
  
\subsubsection{Post-Interaction Phase}
After the experiment, the participant completed a survey on demographics, System Usability Scale~\cite{brooke1996sus}, Perception of Agency~\cite{trafton2023perception}, NASA Task Load Index~\cite{hart2006nasa}, and Trust in Automation~\cite{korber2019theoretical} measures. Finally, the experimenter debriefed the participant about the deception, explaining that the robot was not actually learning from them in real-time and that they would receive the maximum compensation regardless of the robot's performance.


\subsection{Measures}
\label{subsection: measures}

We now define the metrics we use to evaluate each RQ. 

\subsubsection*{RQ1: When do people correct the robot?}
\label{sec:correction timing}

Kullback–Leibler divergence (KLD)~\cite{kullback1951information} has been widely adopted by learning researchers as a metric to model the human's perceived alignment between the robot's understanding of the task
and the actual task constraints~\cite{sadigh2017active, hu2022active, wang2023active}, where a smaller KLD indicates that the human believes the robot has more accurately learned the task. In addition, we included two other timing-related heuristic measures: the time elapsed prior to the correction and the proportion of the robot's intended trajectory that remained untraveled.

\begin{itemize}
\item \textbf{Task Objective Divergence}: We estimate the distribution of plausible motion goals supported by the robot's motion at each timestep~\cite{laidlaw2022boltzmann}. When the participant intervenes to correct the robot's motion, we report the KLD between this distribution (based on the robot's motion thus far) and the actual, correct motion goal. 
\item \textbf{Time Until Correction}: The time (in seconds) between the start of the robot's trajectory and the first correction.
\item \textbf{Proportion of Trajectory Untraveled}: The fraction of the trajectory that was left to be traveled at the time of correction (dashed trajectory in Fig.~\ref{fig:correction}).
\end{itemize}

\subsubsection*{RQ2: How well do people predict robot success/failure?}
\label{sec:intervention accuracy}

Since the presence or absence of human intervention can be construed as a binary label for correct behavior, researchers have tried to leverage this data to train robots to classify correct behavior. We therefore categorize human prediction outcomes within a confusion matrix~\cite{stehman1997selecting}:

\begin{itemize}
\item \textbf{Missed Correction Rate}: The proportion of intended failures that the participant did not correct.
\item \textbf{Unnecessary Correction Rate}: The proportion of intended successes that the participant corrected.
\item \textbf{False Omission Rate}: The proportion of uncorrected trials that were intended failures.
\item \textbf{False Correction Rate}: The proportion of corrected trials that were intended successes.
\end{itemize}

\subsubsection*{RQ3: How do people trade-off precision and effort in their corrections?}

Prior works in LfC have modeled the task objective as a linear combination of relevant features~\cite{bajcsy2017learning, bajcsy2018learning, losey2018including, bobu2020less, bobu2020quantifying, losey2022physical, chisari2022correct, wang2024legibility}, such as goal proximity, object avoidance, and motion smoothness and quantified physical effort using external torque applied by humans on the robot~\cite{bajcsy2017learning, bajcsy2018learning, losey2018including, bobu2020less, bobu2020quantifying, losey2022physical, chisari2022correct, wang2024legibility}.

\begin{itemize}
\item \textbf{Precision}: A linear combination of EEF position and rotation features with respect to the correct goal at the end of the first correction (details in Appendix Sec.~\ref{subsec:precision}).
\item \textbf{Physical Effort}: The L2-norm of the time-integrated torque for the first correction.

\end{itemize}

\section{Results}
\label{sec:results}

For \textbf{RQ1} and \textbf{RQ2}, we conducted a two-way Analysis of Variance (ANOVA)~\cite{fisher1970statistical} with competency and legibility predicting each relevant measure in Section~\ref{subsection: measures}. The main effects were ignored if the interaction effect was
significant. A post-hoc Tukey's honestly significant difference (HSD) test was administered if the legibility main effect or interaction effect was significant. Although it is generally advisable to transform the data so that it follows a normal distribution, Blanca et al. have shown that the ANOVA and F-test are robust to non-normal data~\cite{blanca2017non}. For \textbf{RQ3}, we examined the pairwise condition difference in correlation between task precision and physical effort. We measured statistical significance using a threshold of $\alpha < .05$ and applied the Benjamini–Hochberg procedure for multiple testing corrections~\cite{benjamini1995controlling}. In all figures, asterisks (*) are used to denote statistical significance: a single asterisk indicates $p < 0.05$; double asterisks indicate $p < 0.01$; and triple asterisks indicate $p < 0.001$.

By default, a robot's motion planner will produce predictable (i.e., efficient) trajectories, which we considered as the baseline legibility. Hence, all pairwise comparisons included predictable vs. legible and predictable vs. illegible conditions.

\subsection{Analysis of RQ1}
\label{subsec:correction timing}

\begin{figure*}[ht]
    \centering
    \begin{minipage}{0.325\textwidth}
        \centering
        \includegraphics[width=1.00\textwidth]{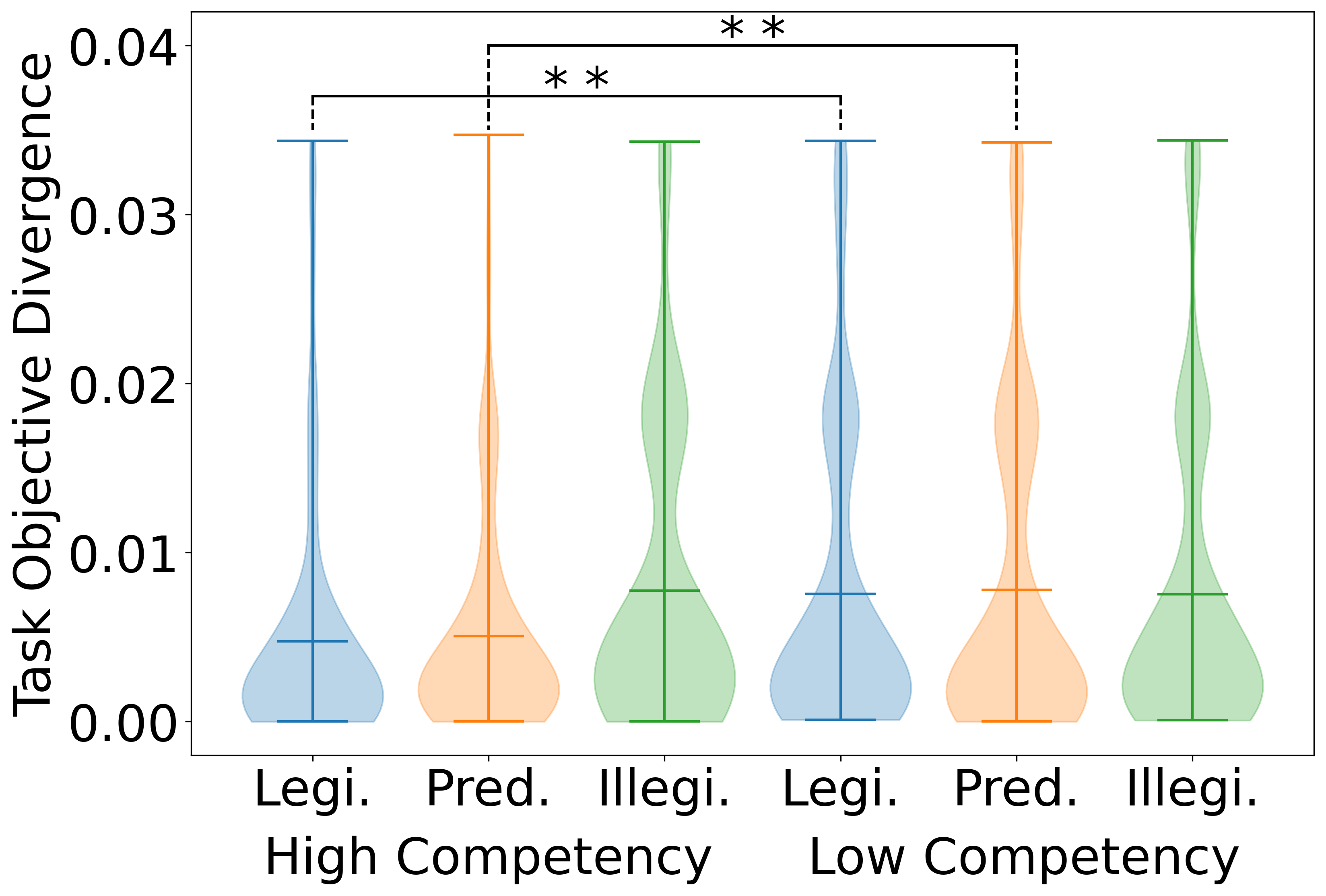}
        \caption{Task Objective Divergence: The KLD between the estimated robot's task belief distribution and the actual, correct motion goal.}
        \label{fig:kld}
    \end{minipage}
    \hfill
    \begin{minipage}{0.325\textwidth}
        \centering
        \includegraphics[width=1.00\textwidth]{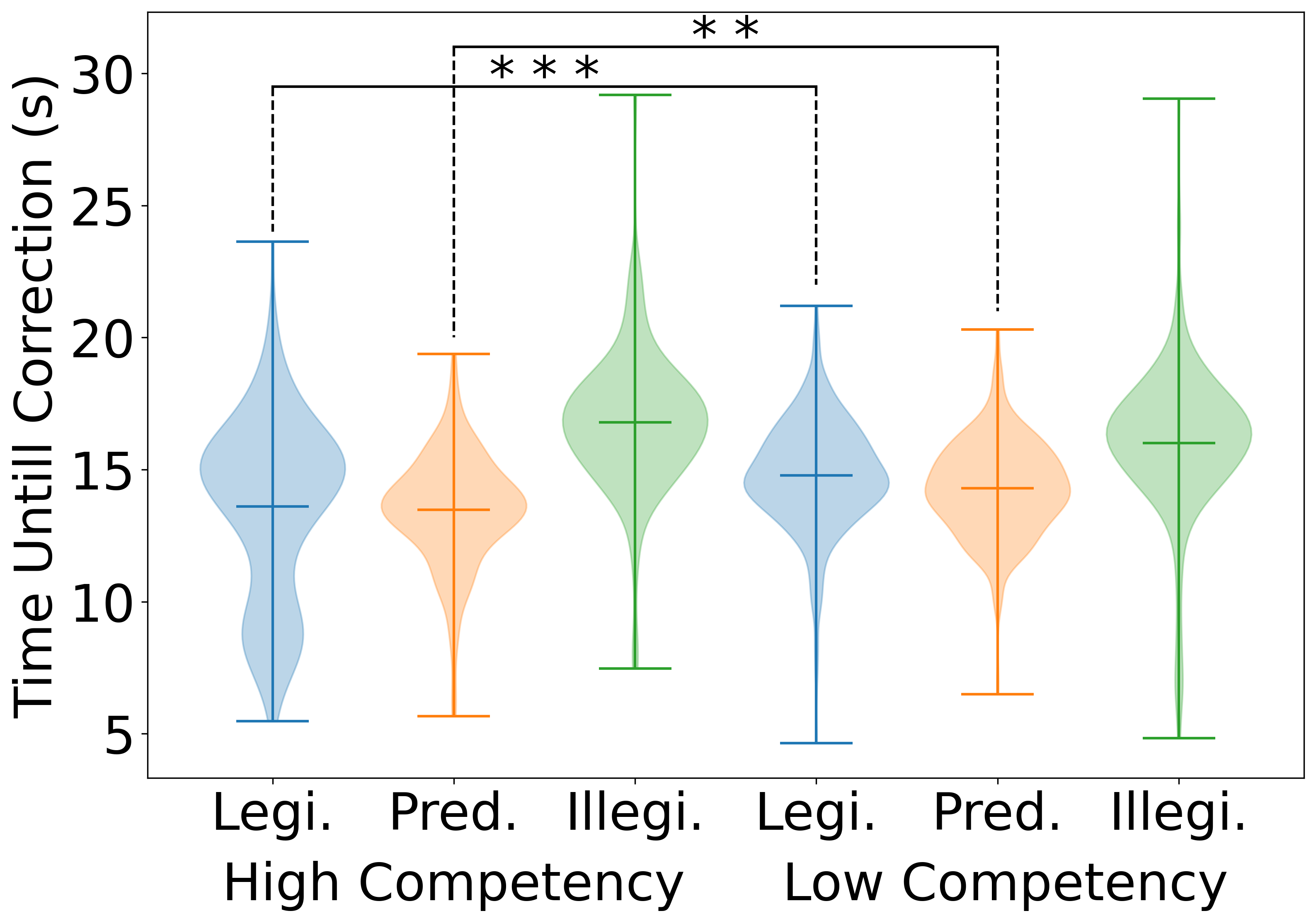}
        \caption{Time Until Correction: The time (measured in seconds) between the start of the robot's trajectory and the first correction.}
        \label{fig:time}
    \end{minipage}
    \hfill
    \begin{minipage}{0.325\textwidth}
        \centering
        \includegraphics[width=1.00\textwidth]{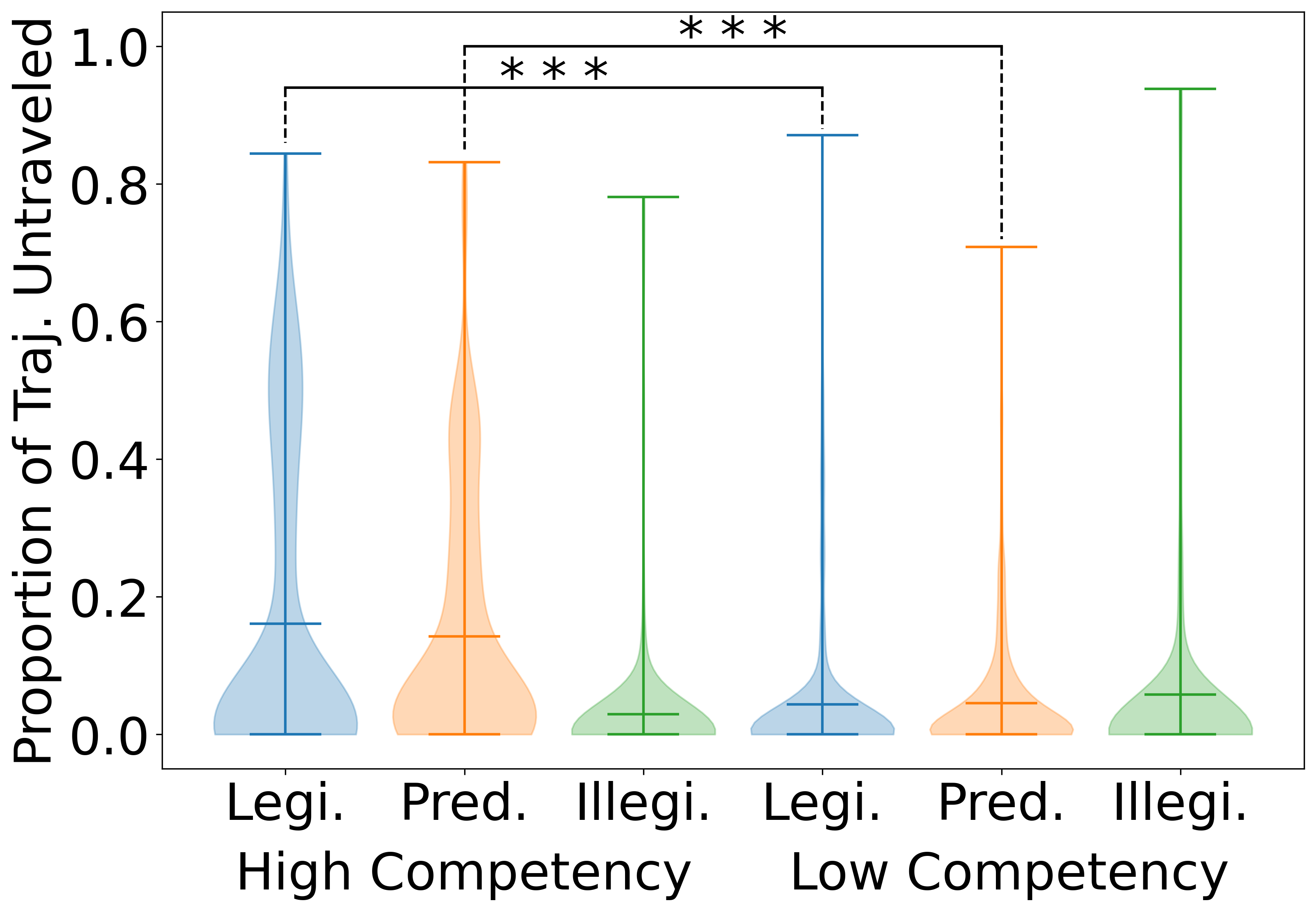}
        \caption{Proportion of Trajectory Untraveled: The fraction of the trajectory that was left to be traveled at the time of correction.}
        \label{fig:traj left}
    \end{minipage}
    \vspace*{-5mm}
\end{figure*}




A two-way ANOVA examining the effect of competency and legibility on task objective divergence showed an interaction effect of 
$(F(2, 1944) = 4.8242, p = 0.0081)$. A Tukey test found that 1) in legible conditions $(p=0.0015)$, people choose to correct the robot when the task objective divergence was smaller in high-competency conditions $(M=0.0047, SD=0.0073)$ than in low competency-conditions $(M=0.0075, SD =0.0090)$, 2) in predictable $(p=0.0055)$ conditions, people choose to correct the robot when the task objective divergence was smaller in high-competency conditions ($M=0.0051, SD=0.0064$) than in low-competency conditions $(M=0.0078, SD=0.0092)$. A two-way ANOVA examining the effect of competency and legibility on time until correction (Fig.~\ref{fig:time}) showed an interaction effect of 
$(F(2, 1944) = 23.3877, p < 0.0001)$. A Tukey's test showed that 1) when motions were legible $(p < 0.0001)$, people correct a competent robot earlier in high-competency conditions $(M=13.5968, SD=3.3740)$ than in low-competency conditions $(M=14.7866, SD=2.0448)$, 2) when motions were predictable $(p=0.0020)$, people correct a competent robot earlier in high-competency conditions $(M=13.4841, SD=2.0831)$ than in low competency-conditions $(M=14.2968, SD=1.7970)$. A two-way ANOVA examining the effect of proportion of trajectory to be traveled (Fig.~\ref{fig:traj left}) showed an interaction effect for competency and legibility of 
$(F(2, 1944) = 34.7386, p < 0.0001)$. A Tukey's test showed that 1) when motions were legible $(p<0.0001)$, robots were corrected later in the trajectory in low-competency $(M=0.0435, SD=0.1136)$ than high-competency conditions $(M=0.1610, SD=0.2265)$ 2) when motions were predictable $(p<0.0001)$, robots were corrected later in the trajectory in in low-competency $(M=0.0457, SD=0.0902)$ than high-competency conditions $(M=0.1422, SD=0.1849)$.

\subsection{Analysis of RQ2}
\label{subsec:label quality}

\begin{figure*}[ht]
    \centering
    \begin{minipage}{0.325\textwidth}
        \centering
        \includegraphics[width=1.00\textwidth]{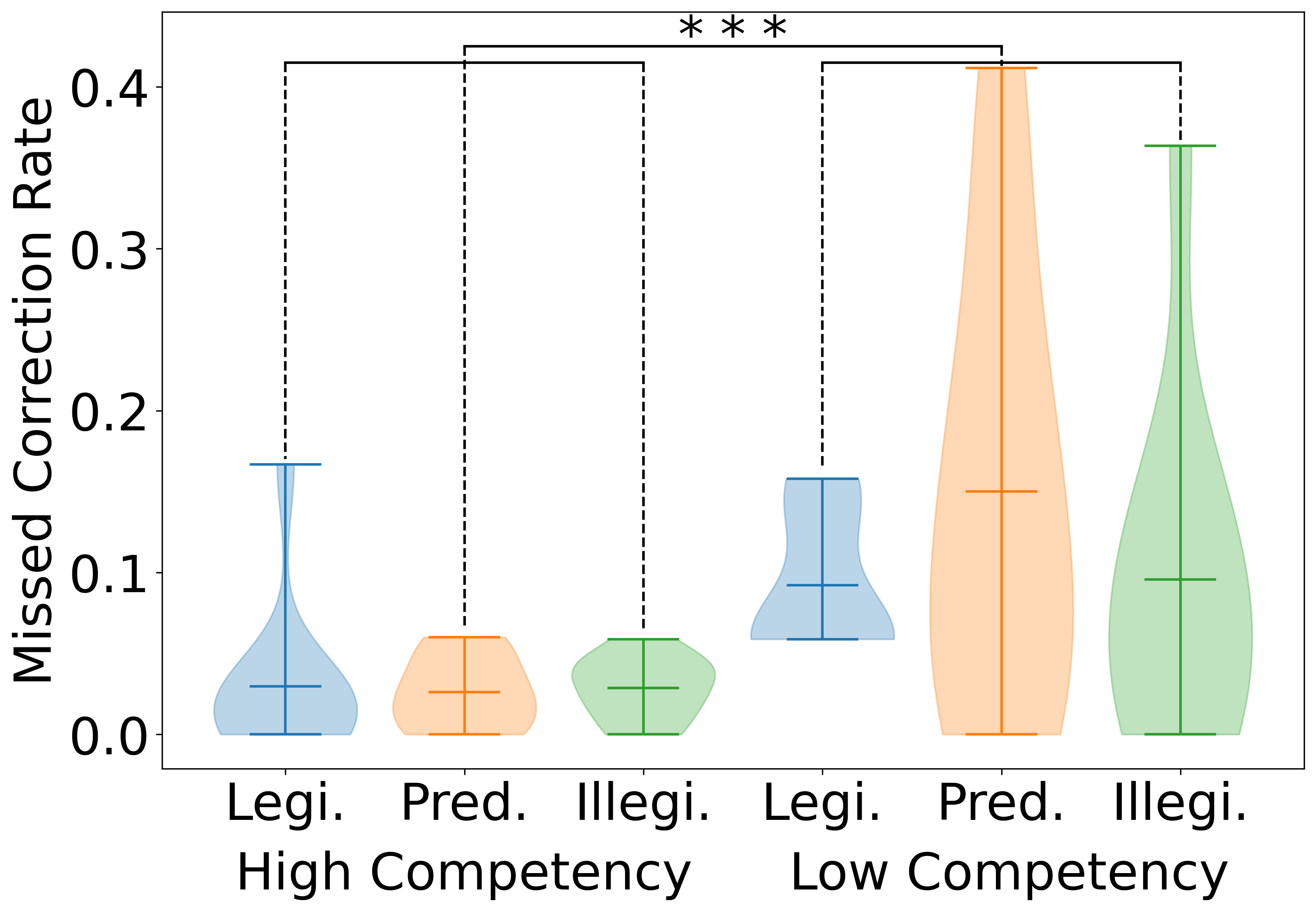}
        \caption{Missed Correction Rate: The proportion of intended failures that the participant did not correct.}
        \label{fig:missed correction rate}
    \end{minipage}
    \hfill
    \begin{minipage}{0.325\textwidth}
        \centering
        \includegraphics[width=1.00\textwidth]{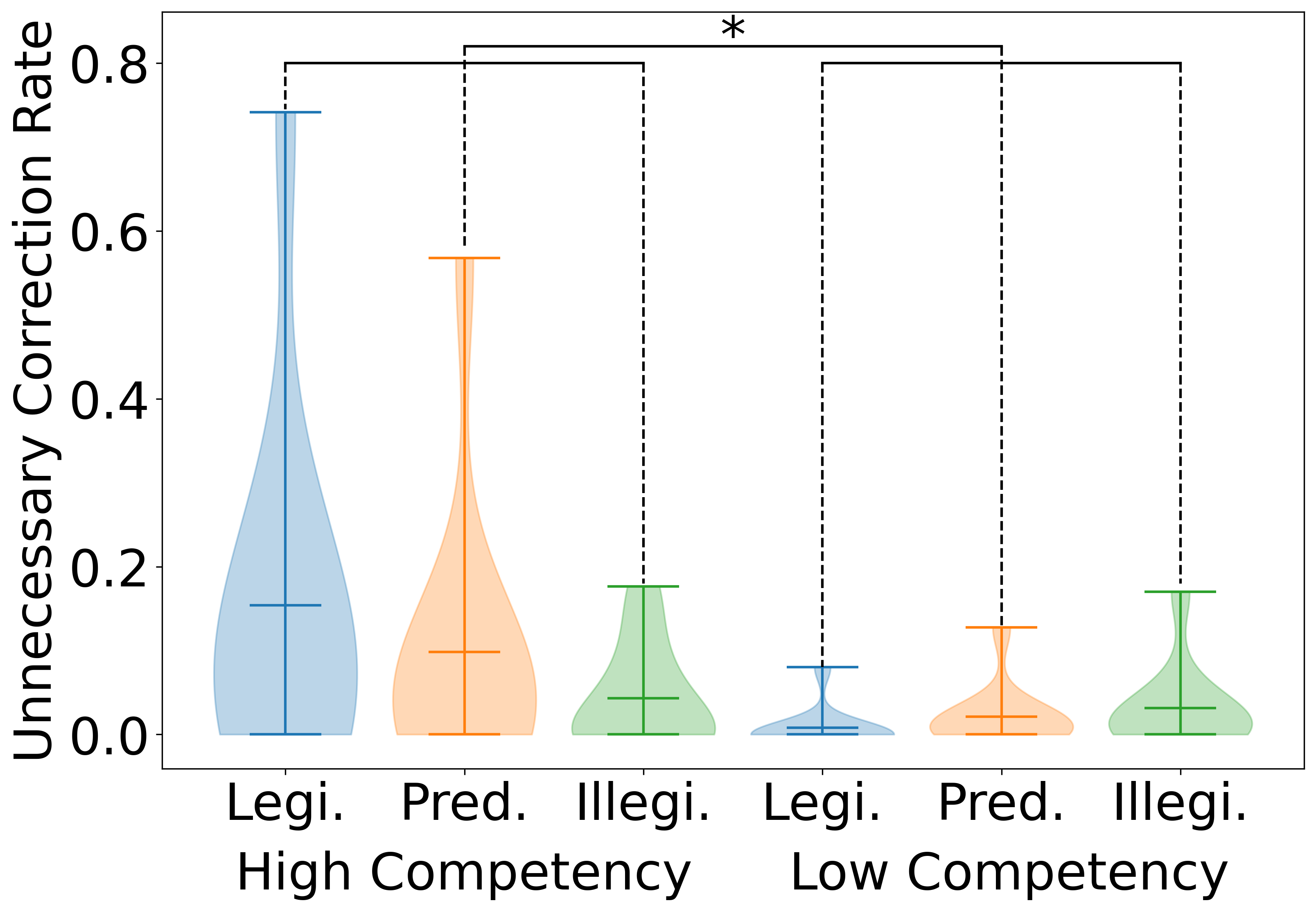}
        \caption{Unnecessary Correction Rate: The proportion of intended successes that the participant corrected.}
        \label{fig:unnecessary correction rate}
    \end{minipage}
    \hfill
    \begin{minipage}{0.325\textwidth}
        \centering
        \includegraphics[width=1.00\textwidth]{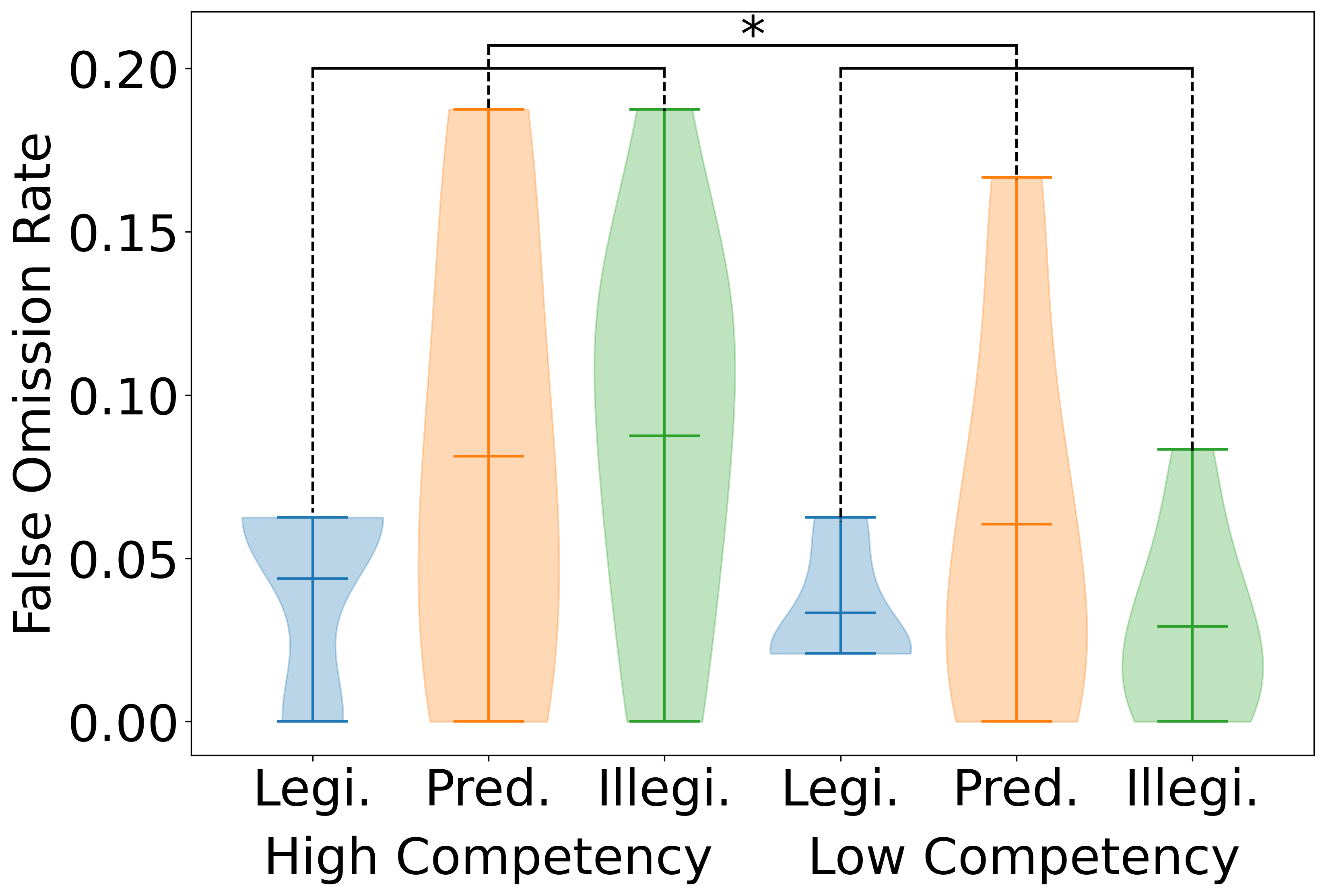}
        \caption{False Omission Rate: The proportion of uncorrected trials that were intended failures.}
        \label{fig:false omission rate}
    \end{minipage}
    \vspace*{-5mm}
\end{figure*}

A two-way ANOVA examining the effect of competency and legibility on missed correction rate (Fig.~\ref{fig:missed correction rate}) showed a competency main effect of 
$(F(1, 54) = 17.4059$, $p < 0.0001)$, indicating that, in low-competency conditions $(M=0.1127, SD=0.1057)$, people were more likely to withhold necessary corrections than in high-competency conditions $(M=0.0281, SD=0.0322)$. A two-way ANOVA examining the effect of competency and legibility on unnecessary correction rate (Fig.~\ref{fig:unnecessary correction rate}) showed a competency main effect of 
$(F(1, 54) = 6.0503$, $p = 0.0171)$, suggesting that, in high-competency conditions $(M=0.0983, SD =0.1695)$, people were more likely to offer unnecessary corrections than in low-competency conditions $(M=0.0202, SD=0.0401)$. From a robot-centric perspective, a two-way ANOVA examining the effect of competency and legibility on false omission rate (Fig.~\ref{fig:false omission rate}) showed a competency main effect of 
$(F(1, 54) = 5.5637$, $p = 0.0220)$, implying that, in high-competency conditions $(M=0.0708, SD=0.0586)$, failures were more likely when corrections were missed than in low-competency conditions $(M=0.0410, SD=0.0404)$. A two-way ANOVA on false omission rate (Fig.~\ref{fig:false correction rate} in the Appendix) showed no significant effect of competency or legibility.

\subsection{Analysis of RQ3: Task and Effort Trade-Off}
\label{subsection:task and effort trade-off}

Two Shapiro-Wilk tests~\cite{shapiro1965analysis} showed that task precision ($W=0.8776$, $p<0.001$) and physical effort ($W=0.7188$, $p<0.001$) were not normally distributed. We, therefore, used the Spearman correlation~\cite{spearman1961proof} to study the relationships between these two variables. Overall, the trend in each condition exhibited a weak positive correlation (Table~\ref{tbl:correlation and sample size} in the Appendix). After adopting a Fisher z-transformation on the correlations~\cite{fisher1915frequency, fisher1921014}, we conducted a pairwise Z-test to examine how competency and legibility affect these positive correlations (Table~\ref{tab:pairwise_comparisons} in the Appendix). We found that, for an incompetent robot with legible motions, the positive correlation of task and effort was significantly weaker compared to an incompetent robot with predictable motions ($p = 0.0075$).

\section{Discussion}
\label{sec:discussion}

Addressing \textbf{RQ1}, we find that, for legible and predictable motions, people chose to intervene as soon as they perceive a minor error in the robot's behavior in high-competency conditions; in low-competency conditions, people were more tolerant of the robot deviating from optimal behavior. This which strongly supports the inverse of \textbf{H1B}; that is, people corrected the robot \emph{earlier} in the trajectory when supervising a highly competent robot (with legible and predictable motions). We find no evidence supporting or disproving \textbf{H1A}.

Corrections are most informative when they reflect the constraints of the task. Therefore, it is advantageous for learning algorithms if people correct the robot only when it is about to make a mistake. To achieve this, robots should avoid deviating from optimal behaviors when highly competent. From a learning perspective, algorithms should not automatically assume that pre-correction trajectories are always informative (particularly for an incompetent robot). Instead, they should primarily leverage the correction trajectory.


For \textbf{RQ2}, we find that competency significantly influences people's ability to accurately predict robot success and failures. 
Our evidence does not support \textbf{H2A} and \textbf{H2B}, but strongly supports their inverse: when supervising an incompetent robot, people are more likely to miss necessary corrections; people are more likely to give corrections when not needed while supervising a highly competent robot.

We find no evidence supporting or disproving \textbf{H2C}; legibility does not significantly affect humans' prediction accuracy. Note that, although both false omissions and false corrections are both incorrect labels, the former are more deleterious than the latter from a learning perspective; false corrections still reflect the correct task constraints, whereas false omissions can cause a robot to mistakenly interpret failures as correct behaviors. Therefore, to improve task learning, it is reasonable to design robot behaviors to minimize missed corrections, even if it also results in people providing more unnecessary ones.







For \textbf{RQ3}, we find that physical effort positively correlates with correction precision, providing empirical evidence to support the assumed trade-off between precision and physical effort (\textbf{H3}). However, when supervising an incompetent robot with legible motions, human corrections exhibited a much weaker correlation than when supervising a robot with predictable motions, hinting that learning researchers should rely less on the heuristics that greater effort suggests more precise task correction. A weaker correlation indicates that extra effort is not necessarily converted to more precise corrections. Therefore, in this condition, solely relying on the task and effort trade-off to model human correction feedback is rather insufficient, and additional features should be explored.

\subsection{Alternatives for Trust-Based Hypotheses}

Several of our hypotheses (\textbf{H1B}, \textbf{H2A}, and \textbf{H2B}) are founded on the idea that people correct the robot based on their trust in it. Yet, we found strong evidence for the \emph{inverse} of these hypotheses. We now propose an alternative explanation for these findings: that people develop high expectations for a highly competent robot, and intervene in the robot's behavior when there is any indication that the robot may not meet this high expectation. On the other hand, people may be giving an incompetent robot the ``benefit of the doubt", waiting until the robot is clearly about to fail before correcting it. Participants being more strict and demanding of highly-competent robots is consistent with previous findings~\cite{paepcke2010judging, belanche2021examining}.

\subsection{Feedback Accuracy Over Time}

To assess the potential effect of people ``calibrating" to the robot's competency level, we repeated the analysis from Sec.~\ref{subsec:label quality}, excluding the first 4, 8, 16, and 32 trials. Further details are provided in Table~\ref{tbl:ablation} in the Appendix. Our analysis confirms the consistency of both human-centric findings across tasks. However, from a robot-centric perspective, no significant differences were observed in whether robots can reliably interpret uncorrected trials as failures later in the experiment, suggesting that this difference is most pronounced when humans are calibrating their expectations of the robot's competency. Additionally, we partitioned the data into the early and late halves of the experiment and conducted a three-way ANOVA on the relevant measures, with the third independent variable representing the data segment (early or late). Results revealed that participants were 
more likely to miss corrections during the early half of the experiment 
compared to the later half
$(F(1, 108) = 6.9635$, $p = 0.0095)$.

\subsection{Takeaway Messages}

We summarize key takeaways for two audiences: interaction designers—who program robot interactions to elicit human feedback best aligned with LfC assumptions—and learning researchers—who develop algorithms to enhance learning outcomes, even when feedback diverges from these assumptions.

\textbf{Implications for Interaction Designers:}
\begin{itemize}
     \item A highly competent robot should avoid deviative behavior, while explorative behavior is more acceptable for a robot that makes mistakes.
     \item Consider how occasionally making deliberate, low-stakes mistakes could reduce human workload from giving frequent, unnecessary corrections.
     \item To incentivize greater effort and better correction precision, consider making motions predictable rather than legible for an incompetent robot.
\end{itemize}

\textbf{Implications for Learning Researchers:}
\begin{itemize}
     \item When designing algorithms for an incompetent robot that leverages pre-correction trajectories to learn task constraints, consider weighting them less (due to their potential for misalignment with the task goal) and assigning a higher weight to the correction trajectory itself.
     \item For active learning researchers, consider allowing more exploratory behavior for a low-competency robot, and less exploration for a highly competent robot.
     \item When interpreting humans' supervision of a highly competent robot, the lack of a correction should be less reliably interpreted as an endorsement of its  behavior.
\end{itemize}



\subsection{Limitations}
\label{subsec:limitations}

Despite collecting over 1,950 correction samples during 3,840 pick-and-place trials, our study's overall sample size remains relatively small (60 participants), with only 10 participants per condition. This limited sample size may have reduced the statistical power for some hypotheses (\textbf{H1A}, \textbf{H2C}) and contributed to the lack of significance observed. For analyses on \textbf{RQ1} and \textbf{RQ3}, we assume each trial data point from each participant as independent to fully leverage the large volume of correction data. However, this assumption may have inflated the degrees of freedom in the statistical tests, as the data points could exhibit high within-participant correlations.


\section{Conclusion}
\label{sec:conclusion}

In this paper, we presented a user study exploring the effects of a robot's competency and motion legibility on how people supervise and correct its behavior. 
We found that when the robot followed predictable or legible motions, people were more sensitive to potential failures by a highly-competent robot compared to an incompetent robot. Additionally, people were more likely to withhold necessary corrections in low-competency conditions and were more prone to offering unnecessary ones in high-competency conditions. Finally, we empirically supported the assumed trade-off between task precision and human effort when giving corrections. However, we highlighted that this trade-off should be less reliably assumed when modeling humans supervising an incompetent robot with legible motions. These findings offer valuable insights for robot interaction designers and learning researchers working with LfC systems and, more broadly, in robot learning scenarios where humans act in a supervisory role.

\section*{Acknowledgments}

This work is partially supported by the National Science Foundation (NSF IIS-2106690, IIS-1955653) and the Office of Naval Research (ONR N00014-24-1-2124). Any opinions, findings, and conclusions or recommendations expressed in this material are those of the authors and do not necessarily reflect the views of the NSF or ONR.

The authors would like to thank Haimanot Belachew, Dražen Brščić, Kate Candon, Jacky Chen, Nicholas C. Georgiou, Ulas Berk Karli, Alexander Lew, Jirachaya ``Fern" Limprayoon, Anushka Potdar, Suba Ramesh, Sydney Thompson, the Yale StatLab, and the Cornell EmPRISE Lab for their feedback and contributions.

\balance

\clearpage
\bibliographystyle{IEEEtran}
\bibliography{refs.bib}
\balance

\clearpage

\section{Appendix}
\label{sec:appendix}

\subsection{Precision}
\label{subsec:precision}
Precision was computed using the EEF position and rotation at the end of the correction. We computed the negative of the EEF position error (Euclidean distance) and rotation error (Quaternion distance) to the correct goal, normalized the two values by their mean and standard deviation, and took their equally weighted linear combination.

\subsection{Supplementary Data}

\begin{figure}[h!]
\centering
\includegraphics[width=0.48\textwidth]{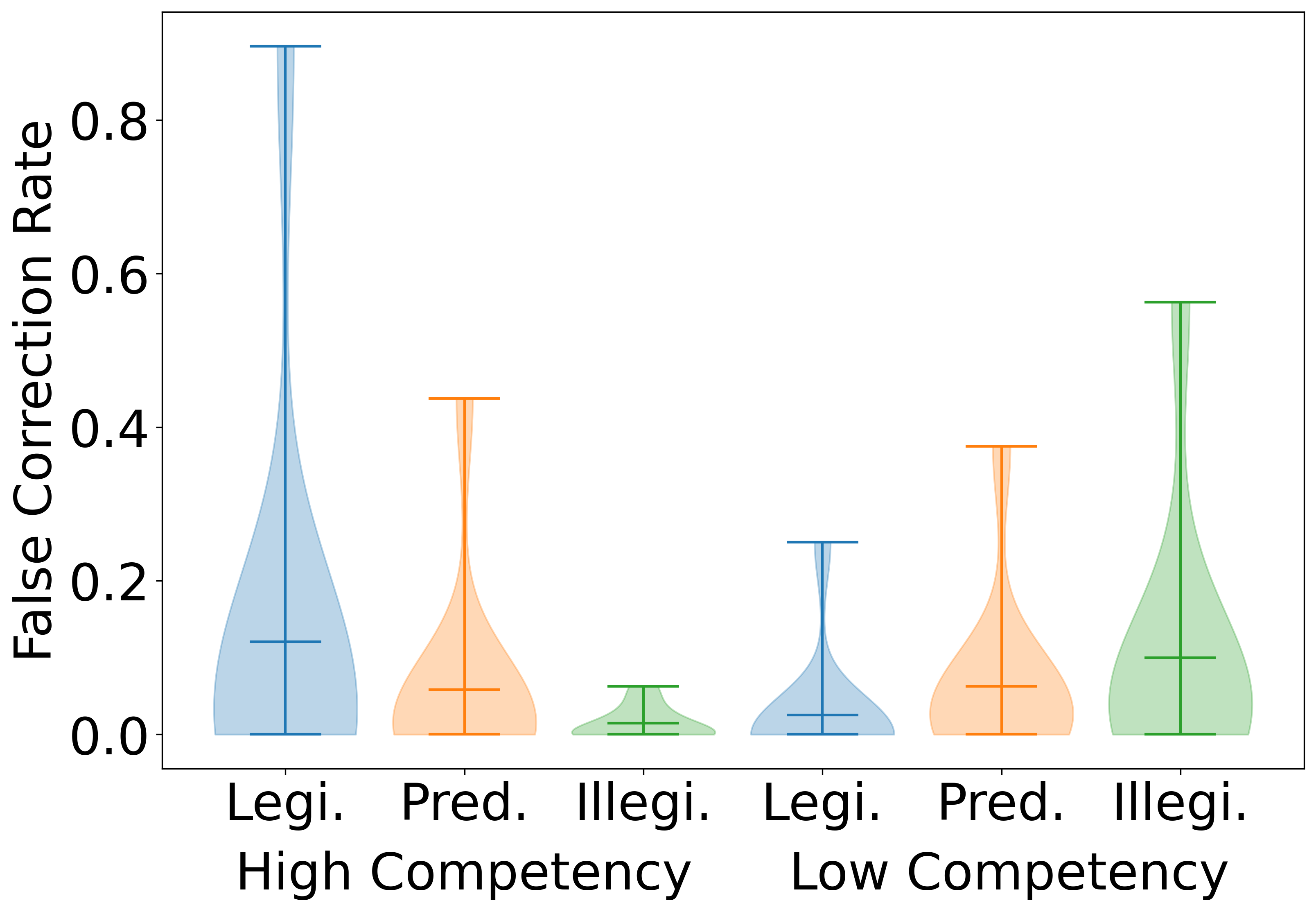}
\caption{False Correction Rate: The proportion of corrected trials that were intended successes.}
\label{fig:false correction rate}
\end{figure}


\begin{table}[h!]
\centering
\begin{tabular}{|c|c|c|}
\hline
   \backslashbox{Legibility}{Competency}  & \textbf{High} & \textbf{Low} \\
\hline
\textbf{Predictable} & 
\begin{tabular}{@{}l@{}}
$\rho = 0.0509$ \\
$n = 175$ \\
\end{tabular} &
\begin{tabular}{@{}l@{}}
$\rho = 0.1840$ \\
$n = 461$ \\
\end{tabular} \\
\hline
\textbf{Illegible} & 
\begin{tabular}{@{}l@{}}
$\rho = 0.0183$ \\
$n = 153$ \\
\end{tabular} &
\begin{tabular}{@{}l@{}}
$\rho = 0.1747$ \\
$n = 482$ \\
\end{tabular} \\
\hline
\textbf{Legible} & 
\begin{tabular}{@{}l@{}}
$\rho = 0.0676$ \\
$n = 211$ \\
\end{tabular} &
\begin{tabular}{@{}l@{}}
$\rho = 0.0101$ \\
$n = 468$ \\
\end{tabular} \\
\hline

\end{tabular}
\vspace{0.1cm} 
\caption{Spearman Correlation Between Task Precision and Physical Effort by Legibility and Competency}
\label{tbl:correlation and sample size}
\end{table}


\begin{table}[h!]
    \centering
    \begin{tabular}{@{}ll
                    S[table-format=1.6,table-align-text-post=false]
                    S[table-format=1.6,table-align-text-post=false]@{}}
        \toprule
        \textbf{Group 1}         & \textbf{Group 2}         & \textbf{$z$-diff} & \textbf{$p$ value} \\ \midrule
        (illegible, high)     & (illegible, low)      & -1.690991      & 0.090839   \\
        (illegible, high)     & (predictable, high)   & -0.292769      & 0.769699   \\
        (illegible, high)     & (predictable, low)    & -1.784440      & 0.074352   \\
        (illegible, low)      & (predictable, high)   &  1.411950      & 0.157965   \\
        (illegible, low)      & (predictable, low)    & -0.147718      & 0.882565   \\
        (legible, high)       & (legible, low)        &  0.690858      & 0.489655   \\
        (legible, high)       & (predictable, high)   &  0.162847      & 0.870639   \\
        (legible, high)       & (predictable, low)    & -1.415818      & 0.156829   \\
        (legible, low)        & (predictable, high)   & -0.457683      & 0.647180   \\
        (legible, low)        & (predictable, low)    & -2.673583      & 0.007505   \\
        (predictable, high)   & (predictable, low)    & -1.511426      & 0.130680   \\
        \bottomrule
    \end{tabular}
    \vspace{0.1cm} 
    \caption{Pairwise Comparisons of Correlations with Fisher's Z-Test}
    \label{tab:pairwise_comparisons}
\end{table}

\begin{table}[H]
\centering
\begin{tabular}{|c|c|c|c|}
\hline
     & \textbf{MCR} & \textbf{UCR} & \textbf{FOR} \\
\hline
\textbf{4} & 
\begin{tabular}{@{}l@{}}
$F(1, 54) = 8.7897$ \\
$p = 0.0045$ \\
$M_{\text{low}}=0.0987$ \\
$ SD_{\text{low}}=0.0211$ \\
$M_{\text{high}}=0.0301$ \\
$ SD_{\text{high}}=0.0111$ \\
\end{tabular} &
\begin{tabular}{@{}l@{}}
$F(1, 54) = 5.2532$\\
$p = 0.0258$ \\
$M_{\text{low}}=0.0199$ \\
$ SD_{\text{low}}=0.0083$ \\
$M_{\text{high}}=0.0921$ \\
$ SD_{\text{high}}=0.0302$ \\
\end{tabular} &
\begin{tabular}{@{}l@{}}
$F(1, 54) = 3.8812$\\
$p = 0.0540$ \\
$M_{\text{low}}=0.0341$ \\
$SD_{\text{low}}=0.0401$ \\
$M_{\text{high}}=0.0613$ \\
$ SD_{\text{high}}=0.06598$ \\
\end{tabular} \\
\hline
\textbf{8} & 
\begin{tabular}{@{}l@{}}
$F(1, 54) = 10.4228$\\
$ p = 0.0021$ \\
$M_{\text{low}}=0.0860$ \\
$ SD_{\text{low}}=0.1269$ \\
$M_{\text{high}}=0.0137$ \\
$ SD_{\text{high}}=0.0225$ \\
\end{tabular} &
\begin{tabular}{@{}l@{}}
$F(1, 54) = 3.5460$\\
$ p = 0.0651$ \\
$M_{\text{low}}=0.0182$ \\
$ SD_{\text{low}}=0.0416$ \\
$M_{\text{high}}=0.0834$ \\
$ SD_{\text{high}}=0.1825$ \\
\end{tabular} &
\begin{tabular}{@{}l@{}}
$F(1, 54) = 0.4369$\\
$ p = 0.5114$ \\
$M_{\text{low}}=0.0307$ \\
$ SD_{\text{low}}=0.0527$ \\
$M_{\text{high}}=0.0404$ \\
$ SD_{\text{high}}=0.0676$ \\
\end{tabular} \\
\hline
\textbf{16} & 
\begin{tabular}{@{}l@{}}
$F(1, 54) = 13.4625$\\
$ p = 0.0006$ \\
$M_{\text{low}}=0.0893$ \\
$ SD_{\text{low}}=0.1110$ \\
$M_{\text{high}}=0.0127$ \\
$ SD_{\text{high}}=0.0183$ \\
\end{tabular} &
\begin{tabular}{@{}l@{}}
$F(1, 54) = 4.4724$\\
$ p = 0.0391$ \\
$M_{\text{low}}=0.0198$ \\
$ SD_{\text{low}}=0.0474$ \\
$M_{\text{high}}=0.0871$ \\
$ SD_{\text{high}}=0.1683$ \\
\end{tabular} &
\begin{tabular}{@{}l@{}}
$F(1, 54) = 0.4553$\\
$ p = 0.5027$ \\
$M_{\text{low}}=0.0306$ \\
$ SD_{\text{low}}=0.0374$ \\
$M_{\text{high}}=0.0389$ \\
$ SD_{\text{high}}=0.0568$ \\
\end{tabular} \\
\hline
\textbf{32} & 
\begin{tabular}{@{}l@{}}
$F(1, 54) = 10.6417$\\
$ p = 0.0019$ \\ 
$M_{\text{low}}=0.0818$ \\
$ SD_{\text{low}}=0.1212$ \\
$M_{\text{high}}=0.0080$ \\
$ SD_{\text{high}}=0.0192$ \\
\end{tabular} &
\begin{tabular}{@{}l@{}}
$F(1, 54) = 4.5391$\\
$p = 0.0377$ \\
$M_{\text{low}}=0.0208$ \\
$ SD_{\text{low}}=0.0470$ \\
$M_{\text{high}}=0.0909$ \\
$ SD_{\text{high}}=0.1746$ \\
\end{tabular} &
\begin{tabular}{@{}l@{}}
$F(1, 54) = 0.1644$\\
$p = 0.6868$ \\
$M_{\text{low}}=0.0306$ \\
$ SD_{\text{low}}=0.0082$ \\
$M_{\text{high}}=0.0250$ \\
$SD_{\text{high}}=0.0111$ \\
\end{tabular} \\
\hline
\end{tabular}
\vspace{0.1cm} 
\caption{Results from the Same Analysis in Sec.~\ref{subsec:label quality} (Two-Way ANOVA) But Removing the First 4, 8, 16, and 32 Trials of the Data. Values in the first column indicate the number of trials removed from the analysis. MCR stands for missed correction rate. UCR stands for unnecessary correction rate. FOR stands for false omission rate.}
\label{tbl:ablation}
\end{table}

\end{document}